\documentclass[conference]{IEEEtran}
\IEEEoverridecommandlockouts
\usepackage{amsmath,url}

\usepackage{cite}
\usepackage{amsmath,amssymb,amsfonts, bm}
\DeclareMathOperator*{\argmin}{argmin}
\newcommand{\R}{\mathbb{R}}
\usepackage{enumerate}
\usepackage[linesnumbered,ruled,vlined]{algorithm2e}
\SetKwComment{Comment}{$\triangleright$\ }{}
\let\oldnl\nl% Store \nl in \oldnl
\newcommand{\nonl}{\renewcommand{\nl}{\let\nl\oldnl}}% Remove line number for one line
\newcommand{\vars}{\texttt}
\usepackage[]{graphicx}
\graphicspath{{Figures/}}
\usepackage{support-caption}
\usepackage{subcaption}
\usepackage{lipsum}
\usepackage{balance}
\usepackage{textcomp}
\usepackage{xcolor}
\usepackage{multirow}
\usepackage{color}
\usepackage{listings}
\usepackage{courier}
\usepackage{float}
\usepackage{verbatim}
\usepackage{array}
\usepackage{fancyhdr}
\definecolor{CustomColor}{RGB}{255, 255,255}

\lstdefinelanguage{turtle}
{
    columns=fullflexible,
    keywordstyle=\color{red},
    morekeywords={@prefix,@base,@forSome,@forAll,@keywords},
    morecomment=[l]{\#},
    tabsize=4, 
    alsoletter={-?}, % allowed in names
    morecomment=[s][\color{blue}]{<}{>},
    basicstyle=\ttfamily\color{black}, 
    morestring=[b][\color{black}]\",    
    backgroundcolor=\color{CustomColor}
}
\lstdefinestyle{turtle}{%
    morekeywords={a, @prefix},
    morecomment=[s][\rmfamily]{<}{>},
    morecomment=[s][\itshape]{"}{"},
}

\def\BibTeX{{\rm B\kern-.05em{\sc i\kern-.025em b}\kern-.08em
    T\kern-.1667em\lower.7ex\hbox{E}\kern-.125emX}}

\usepackage{nomencl}
\makenomenclature

\begin{document}
\title{A Clustering Framework for Residential \\Electric Demand Profiles\thanks{The  ADAPT  Centre  for  Digital  Content  Technology  is  funded  under  the SFI Research Centres Programme (Grant 13/RC/2106) and is co-funded under the European Regional Development Fund.}}

\author{\IEEEauthorblockN{Mayank~Jain}
\IEEEauthorblockA{\textit{ADAPT SFI Research Centre} \\
\textit{University College Dublin (UCD)}\\
Dublin, Ireland\\
mayank.jain@adaptcentre.ie}
\and
\IEEEauthorblockN{Tarek~AlSkaif}
\IEEEauthorblockA{\textit{Information Technology Group} \\
\textit{Wageningen University and Research}\\
Wageningen, The Netherlands\\
tarek.alskaif@wur.nl}
\and
\IEEEauthorblockN{Soumyabrata~Dev}
\IEEEauthorblockA{\textit{ADAPT SFI Research Centre} \\
\textit{University College Dublin (UCD)}\\
Dublin, Ireland\\
soumyabrata.dev@adaptcentre.ie}
}

\maketitle

\begin{abstract}
The availability of residential electric demand profiles data, enabled by the large-scale deployment of smart metering infrastructure, has made it possible to perform more accurate analysis of electricity consumption patterns. This paper analyses the electric demand profiles of individual households located in the city Amsterdam, the Netherlands. A comprehensive clustering framework is defined to classify households based on their electricity consumption pattern. This framework consists of two main steps, namely a dimensionality reduction step of input electricity consumption data, followed by an unsupervised clustering algorithm of the reduced subspace. While any algorithm, which has been used in the literature for the aforementioned clustering task, can be used for the corresponding step, the more important question is to deduce which particular combination of algorithms is the best for a given dataset and a clustering task. This question is addressed in this paper by proposing a novel objective validation strategy, whose recommendations are then cross-verified by performing subjective validation.
\end{abstract}


% Nomenclature
\nomenclature{$n$}{Number of households}
\nomenclature{$r$}{Hourly resolution of original data}
\nomenclature{$d'$}{Optimal number of reduced dimensions}
\nomenclature{$k$}{Optimal number of clusters}

\printnomenclature

\section{Introduction}
\label{sec:intro}

In recent years, a large increase in the number of installed smart metering systems has taken place world-wide. In Europe, this has been triggered by the European Union (EU) regulations and directives on the deployment of smart metering systems. This is also in line with the urgent need of phasing out non-renewable energy resources~\cite{monasterolo2019impact}, and provide better and cleaner energy alternatives to citizens. For instance, the EU Recommendation 2012/148/EU deeply focused on the roll-out of smart metering infrastructure, stating that at least 80\% of electricity meters in EU member state buildings will be replaced by smart meters by the end of 2020~\cite{Council2012}.

A smart meter is a digital electric meter that measures and records the consumption of electric power in buildings during short time intervals (\textit{e.g.} every 10 s), and communicates that through a two-way communication system, to the energy supplier for monitoring and billing processes. Sampling times of extremely short intervals are rarely implemented in practice, due to the huge consumption of digital and communication resources. However, these smart meters collected at moderately spaced time intervals generate vast amount of data, analysis of which could enable energy suppliers to better understand their customers' energy consumption behavior~\cite{alskaif2018gamification}. It could also provide invaluable insights for grid operators about grid status in real time and enable advanced power management solutions~\cite{brinkel2020impact}.

Importance of analyzing energy consumption patterns in residential areas is widely recognized and has attracted many researchers to work in the area%~\cite{hayn2014}. 
~\cite{alskaif2018gamification}. Clustering households based on their load demand profiles is one of the key tasks relevant to this research which can be performed on the dataset obtained from smart meters. While numerous methods to perform clustering have been proposed in the literature (more details in~\ref{sec:litReview}), the bigger question is to comment on the validity and correctness of the results obtained. In this work, we propose a novel scheme to perform cluster validation which is especially suited for the task of clustering residential electric demand profiles.

\subsection{Relevant Literature} \label{sec:litReview}
Several clustering frameworks have been proposed in literature so far. Most of these are comprised of a dimensionality reduction algorithm followed up by a clustering algorithm~\cite{al2016,yildiz2018}. Clustering validation indices (CVIs), like silhouette score, Davies-Bouldin score, and Calinski-Harabasz score, are most commonly used to perform validation of a clustering algorithm~\cite{yilmaz2019comparison}. However, different indices often lead to different results which further complicate the task~\cite{Chicco2012}. Another method which has been proved very effective is to perform manual inspection of the obtained clusters and check if the results are acceptable or not~\cite{yildiz2018, Chicco2012}.

\subsection{Contributions \& Organization of the Paper}
The contribution of this paper can be summarized as follows:

\begin{itemize}
    \item A generalized two-step framework for clustering residential electric demand profiles has been defined. Most methods to perform a clustering task can be defined in the same framework.
    \item A novel objective validation strategy has been proposed to validate the clustering results. Moreover, its recommendations are cross-verified by performing subjective validation (or, manual inspection).
\end{itemize}

The paper\footnote{With the spirit of reproducible research, the code to reproduce the simulations in this paper is shared at \url{https://github.com/jain15mayank/Clustering-Framework-for-Resedential-Demand-Profiles}} proceeds by describing the dataset along with the pre-processing steps which were undertaken in Section~\ref{sec:dataset}. This is followed by defining the clustering framework and the algorithms which were implemented underneath it (Section~\ref{sec:clusteringFramework}). Section~\ref{sec:results} elaborates on the validation strategies (both objective and subjective) and the results obtained. The paper finally concludes in Section~\ref{sec:Conc} with some comments on the possible extensions of the work.

\section{Dataset and Case Study}
\label{sec:dataset}
The dataset used in this work describes hourly electricity demand of individual households situated in Amsterdam, the Netherlands. The data is collected from $27$ households for five-quarter years duration, from January $2018$ till March $2019$. These households are part of an energy community which was initiated as a pilot during the PARENT project \cite{PARENT}. The PARENT project stands for ``PARticipatory platform for sustainable ENergy managemenT''. It aimed to provide communities with technology and support to help reducing energy consumption in their homes and to investigate ways in which communities can work towards more sustainable lifestyles.

In a live dataset as such, a task as simple as clustering the households based on their demand profiles becomes very challenging. The task is further complicated in this case as most of the data is not available. This is because some households joined in late while others took a premature exit. This challenge makes this work more important as commenting on the correct combination of algorithms for clustering becomes more difficult.

\subsection{Consumption pattern of households}

Although the electricity consumption of households varies largely with seasons~\cite{conevska2020}, this factor cannot be considered in this work as the dataset is small, has missing values and is from a limited time frame of just above $1$ year. However, it is relatively safer to assume that despite of seasonal differences throughout the year, the hourly consumption trend remains same for a particular household - enough to the extent that the household will fall in the same cluster every time (provided the clustering is righteously done). It must be noted that this assumption can be easily removed if enough data is available. Furthermore, this assumption does not create any impact on either the clustering framework or the proposed objective validation strategy.

\begin{figure}[htb]
\begin{subfigure}{.49\textwidth}
  \centering
  % include first image
  \includegraphics[width=.9\linewidth]{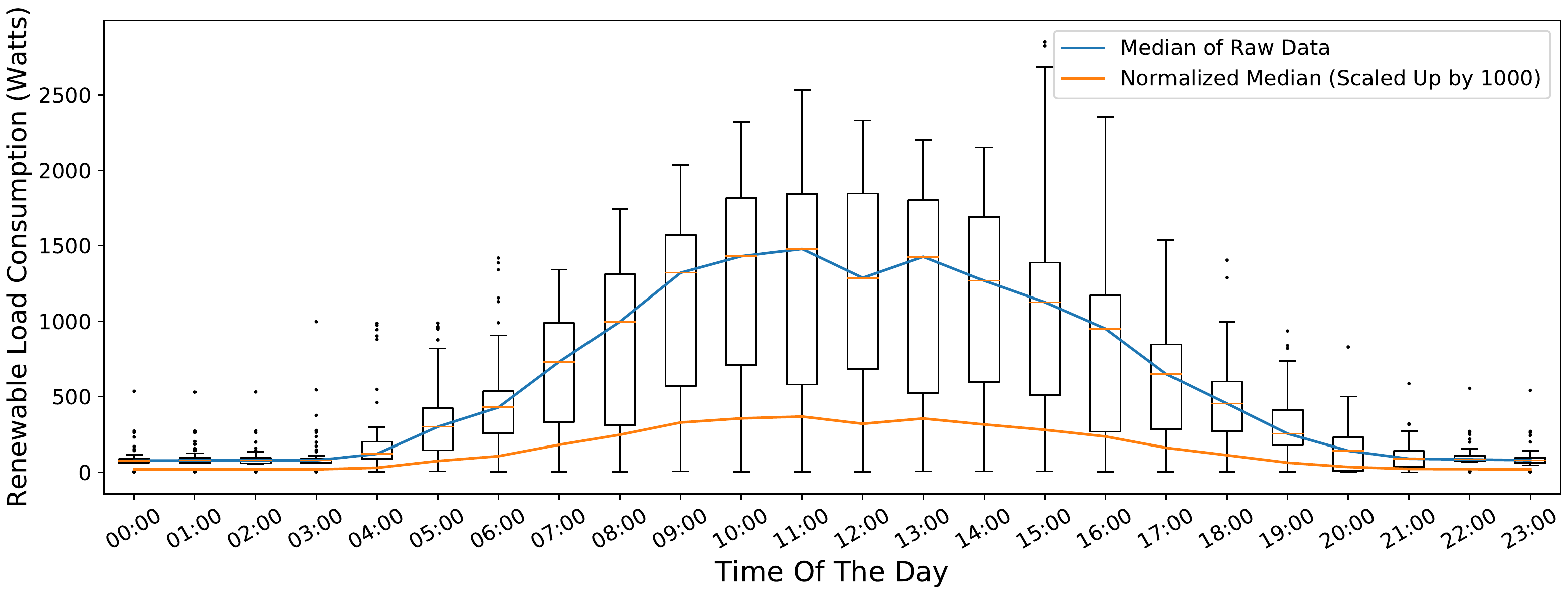}  
  %\caption{Put your sub-caption here}
  \label{fig:box-plot1}
\end{subfigure}\vspace{3pt}
\newline
\begin{subfigure}{.49\textwidth}
  \centering
  % include third image
  \includegraphics[width=.9\linewidth]{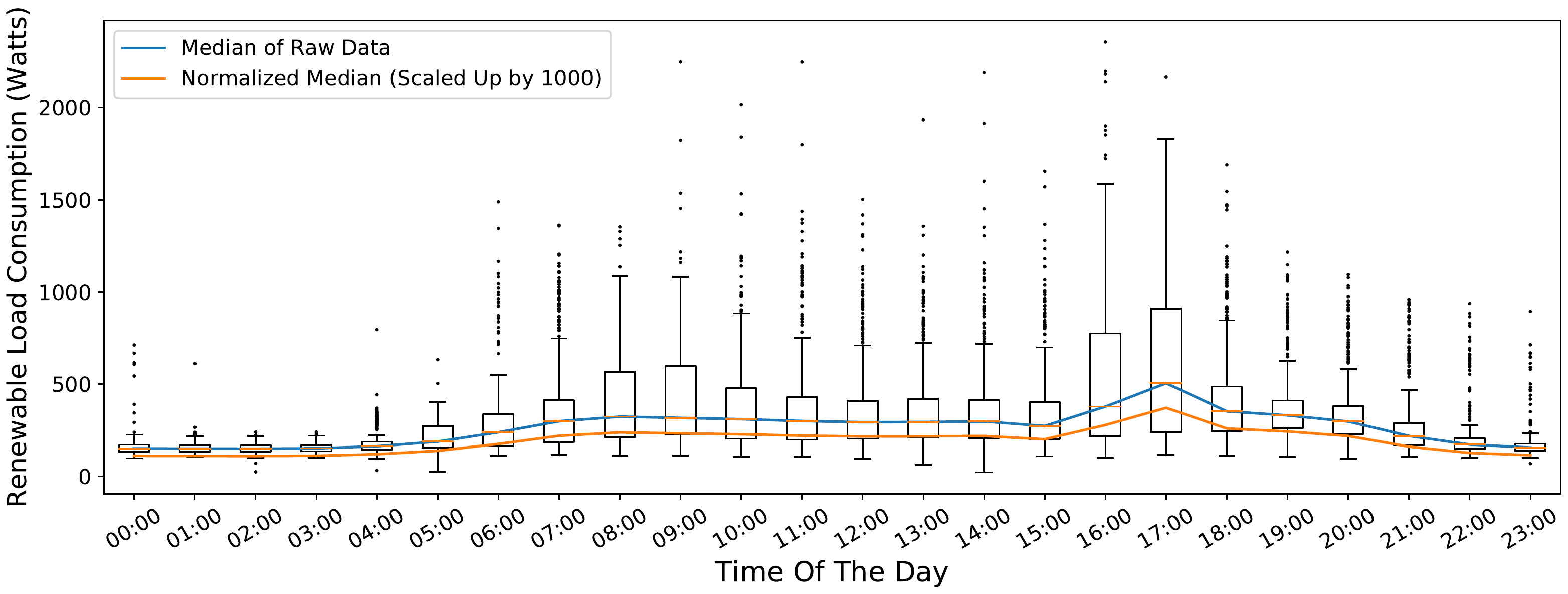}  
  %\caption{Put your sub-caption here}
  \label{fig:box-plot3}
\end{subfigure}
\caption{Box-plots of daily consumption pattern for $2$ different households, also depicting the median and the scaled up version of normalized median.}
\label{fig:BoxPlots}
\end{figure}

Figure~\ref{fig:BoxPlots} shows the box-plots for daily consumption trends for $2$ different households. Due to a large number of outliers in the daily consumption trend, mean cannot be selected as a good measure of central tendency as it will become heavily biased. The mode value will be inappropriate as well since the dataset is too small. However, using the median value will assist us in effectively ignoring the contribution of the outliers, and focus on understanding the general consumption pattern of electric load. Therefore, median was selected as the measure of central tendency in this case.

\subsection{Pre-processing}
To identify consumption trends, it is required to normalize the median curve obtained for each household. This is done by the standard $\ell_2$-normalization. The complete pre-processing steps are defined in Algorithm~\ref{algo:preProcess}.

\begin{algorithm}
%\DontPrintSemicolon % Some LaTeX compilers require you to use \dontprintsemicolon instead
$\vars{n} \gets \text{number of households}$\\
$\vars{r} \gets \text{hourly resolution of original data}$\\
$\vars{d} \gets 24/r$ \text{(dimensionality)}\\
$\bm{M_{n \times d}} \gets$ median daily consumption of each household\linebreak stored row-wise\\
$\bm{M'_{n \times d}} \gets \ell_2\text{-Normalization(}\bm{M}\text{, row-wise)}$\\
\textbf{return} $\bm{M'}$
\caption{Pre-Processing the Dataset}
\label{algo:preProcess}
\end{algorithm}

\section{Clustering Framework}
\label{sec:clusteringFramework}

\begin{figure}[!ht]
    \centering
    \includegraphics[width=0.45\textwidth]{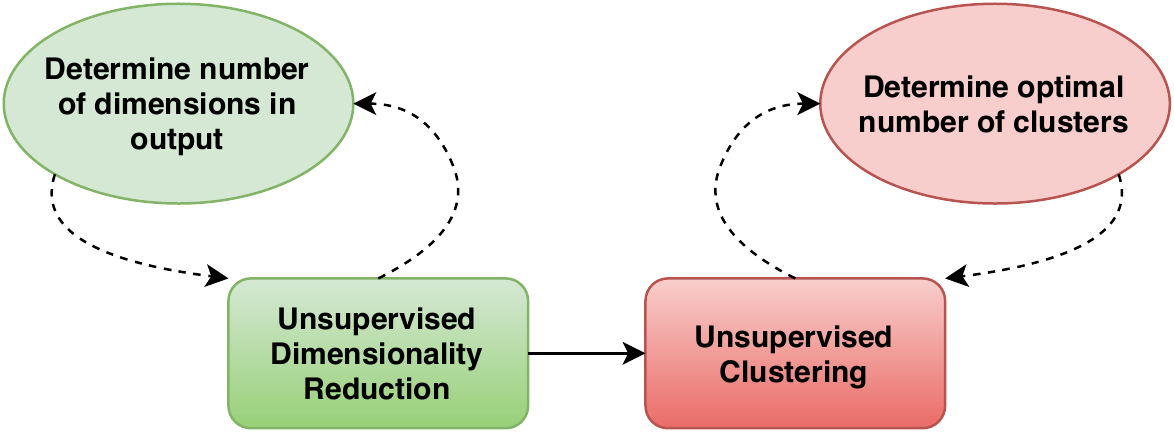}
    \caption{Overview of the clustering framework.}
    \label{fig:cluster_framework}
\end{figure}

To achieve meaningful clustering of a dataset in higher dimensions, dimensionality reduction is generally a pre-requisite~\cite{gonzalez2019condition}. The primary aim of a good dimensionality reduction technique is to capture most information in lower dimension subspace so that the \textit{distance} between $2$ points carries more meaning in terms of the similarity. In this regard, a generalized clustering framework can broadly be defined in two major steps, \textit{i.e.} dimensionality reduction followed by the unsupervised clustering algorithm (\textit{cf.} Fig.~\ref{fig:cluster_framework}). However, both of these steps have an intermediate stage. In the case of dimensionality reduction step, the task is to find the optimal number of dimensions to which the original data should be reduced to such that the loss of information is minimal~\cite{manandhar2018systematic}. Similarly, for the clustering algorithm, there is a requirement to find the optimal number of clusters.

The literature~\cite{ciarelli2009, alskaif2020systematic} offers many algorithms for both unsupervised dimensionality reduction and unsupervised clustering. The suitability of these algorithms depends upon the task and data at hand. This also follows from the `no free lunch' theorem~\cite{wolpert2002}. Although, any combination can be tried and tested for a particular task, this paper implements only a few to provide a good platform for testing the proposed objective validation strategy (\emph{cf.} Section~\ref{sec:objectiveValidation}). The $2$ algorithms are implemented for dimensionality reduction, \textit{viz.} Feature Agglomeration (FA) and Principal Component Analysis (PCA). The FA is a modified version of agglomerative clustering where features are clustered instead of samples. Likewise, $2$ algorithms are implemented for clustering, namely, Spectral Clustering (SC), and K-Means Clustering (KMC). This leads to a total of $4$ different combinations as per our clustering framework.

\subsection{Hyper-parameter settings}
As discussed before, both the steps in the clustering framework require an intermediate step to determine the optimal number of reduced dimensions (for dimensionality reduction step) and the optimal number of clusters (for clustering step). These steps are popularly referred as the hyper-parameter settings in the literature~\cite{yildiz2018,jain2020forecasting}.

\begin{figure}[htb]
\begin{subfigure}{.47\textwidth}
  \centering
  % include first image
  \includegraphics[width=.8\linewidth]{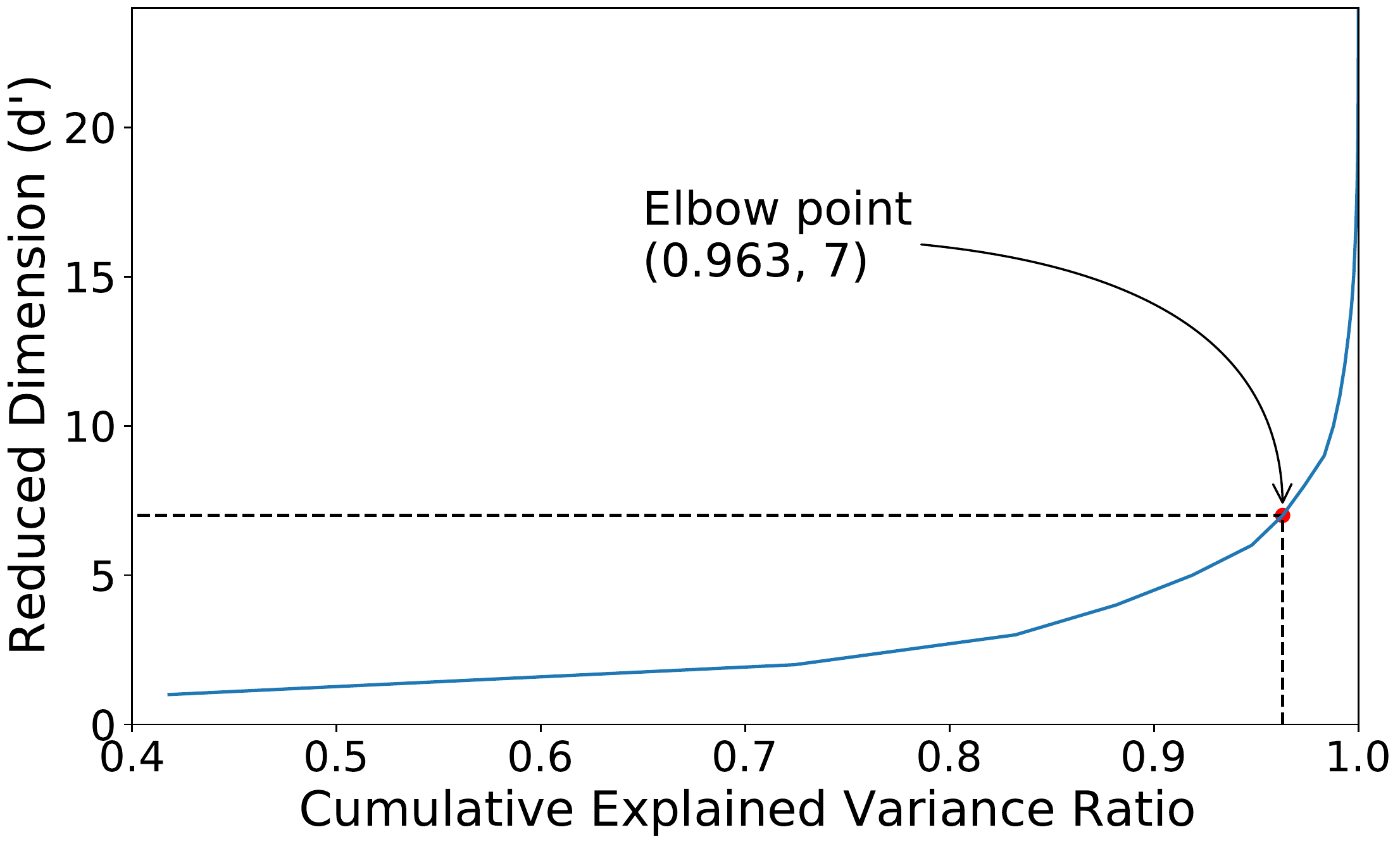}  
  \caption{CEVR vs. $d'$ for PCA hyper-parameter setting.}
  \label{fig:elbow-pca}
\end{subfigure}\vspace{4pt}
\newline
\begin{subfigure}{.47\textwidth}
  \centering
  % include second image
  \includegraphics[width=.8\linewidth]{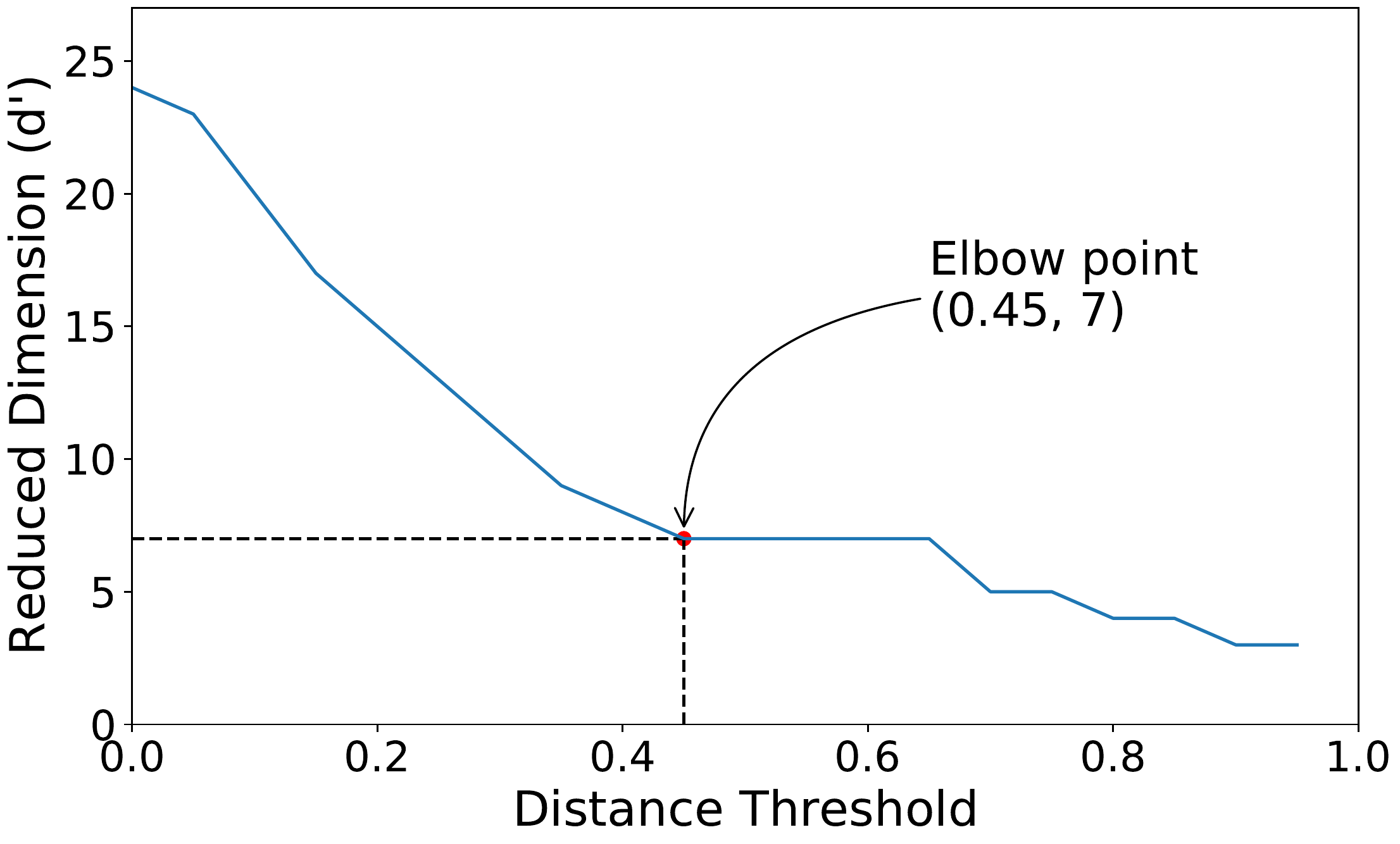}  
  \caption{$thr_{FA}$ vs. $d'$ for FA hyper-parameter setting.}
  \label{fig:elbow-fa}
\end{subfigure}
\caption{Elbow heuristics applied to obtain the value of CEVR (in Fig.~\ref{fig:elbow-pca}) and $thr_{fa}$ (in Fig.~\ref{fig:elbow-fa}) by plotting different value of these thresholds against the corresponding reduced number of dimensions.}
\label{fig:dimRedHP}
\end{figure}

For both the dimensionality reduction algorithms, the optimal number of reduced dimensions (say, $d'$) were determined by the elbow heuristics~\cite{onset2016}. Technically, PCA only calculates the eigenvectors of the covariance matrix of the original data and thus, every corresponding eigenvalue defines the explained variance by the eigenvector. If explained variance ratio (EVR) is defined by the $\ell_1$ normalization of eigenvalues, cumulative explained variance ratio (CEVR), corresponding to a particular eigenvalue-eigenvector pair, is the cumulative EVR of all pairs whose eigenvalue is greater than the pair under consideration, and the pair itself~\cite{manandhar2019data}. Furthermore, the PCA reduces the dimensions by selecting the top $d'$ eigenvector candidates based on CEVR values. Hence, the graph between CEVR and $d'$ is drawn on which elbow-heuristics is performed to determine the final values of hyper-parameters: `CEVR' $= 0.96$, and $d'_{PCA} = 7$ (\textit{cf.} Fig.~\ref{fig:elbow-pca}). Similarly, there is a factor called distance threshold (say, $thr_{FA}$) in the case of feature agglomeration~\cite{daniels2006}. As the value of $thr_{FA}$ increases, $d'$ decreases, and more information is lost in the process. In this case, elbow heuristics is performed on the graph between $thr_{FA}$ and $d'$ to determine the final values of hyper-parameters: $thr_{FA} = 0.45$, and $d'_{FA} = 7$ (\textit{cf.} Fig.~\ref{fig:elbow-fa}).

\begin{figure}[htb]
\begin{subfigure}{.49\textwidth}
  \centering
  % include first image
  \includegraphics[width=.8\linewidth]{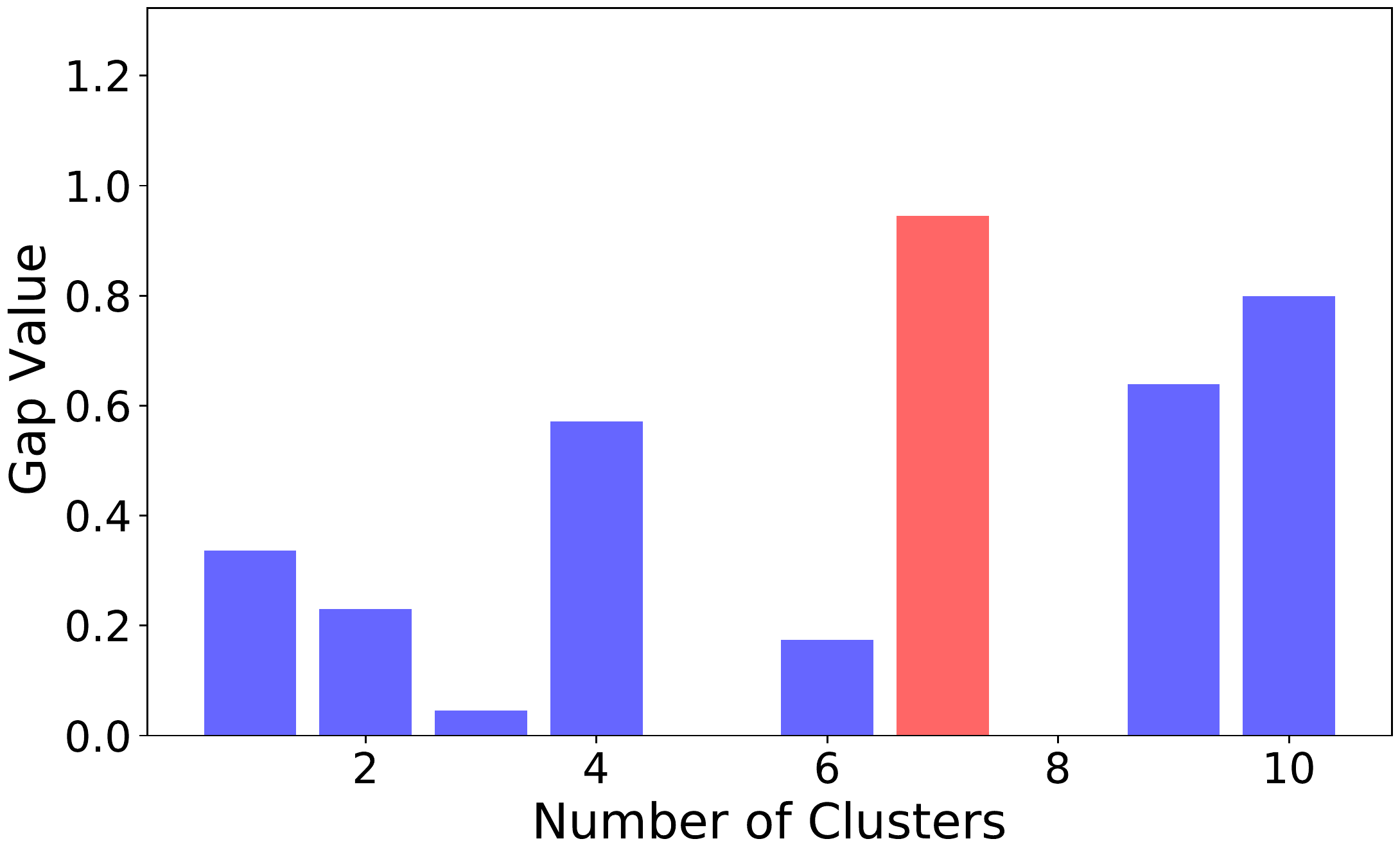}  
  \caption{Gap statistics for FA$+$SC framework.}
  \label{fig:gap-fa-sc}
\end{subfigure}
\newline
\begin{subfigure}{.49\textwidth}
  \centering
  % include second image
  \includegraphics[width=.8\linewidth]{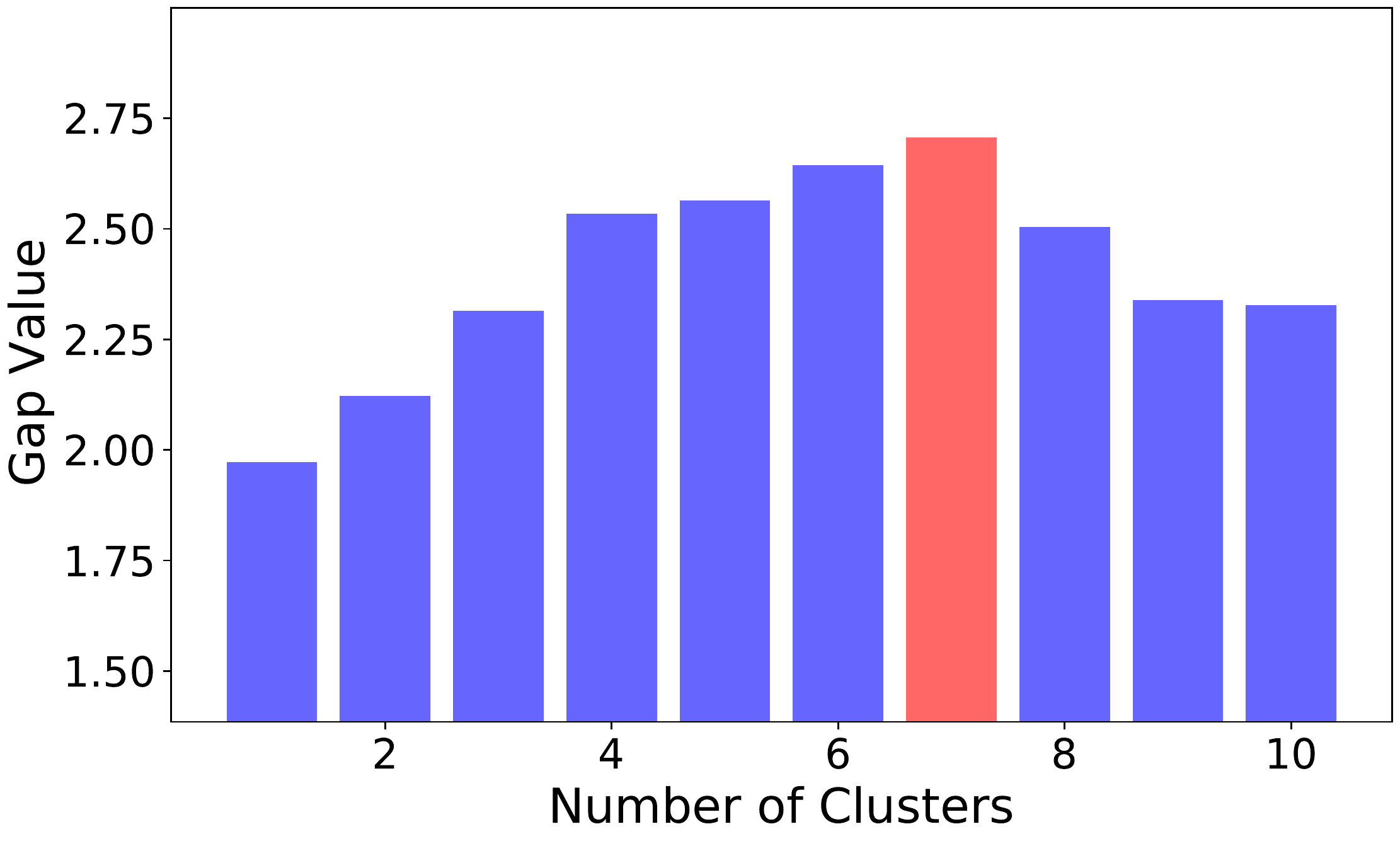}  
  \caption{Gap statistics for PCA$+$KMC framework.}
  \label{fig:gap-pca-km}
\end{subfigure}\vspace{4pt}
\caption{Results of gap statistics as obtained for two different clustering frameworks.}
\vspace{-0.5cm}
\label{fig:clusteringHP}
\end{figure}

Once the dimensions are reduced, gap statistics~\cite{tibshirani2001} is used to find the optimal number of clusters (say, $k$) for both spectral clustering and k-means clustering algorithms. In this process, a statistics called \textit{gap} is calculated corresponding to a given number of cluster. A large value of \textit{gap} would signify that the current clustering distribution is far away from a random uniform distribution of points, and hence depicts better clustering. Accordingly, the number of clusters corresponding to the highest value of \textit{gap}, are selected. It is to be noted that this step is dependent on the preceding step of dimensionality reduction and the process needs to be repeated if anything (algorithm and/or hyper-parameter settings) changes in the preceding step. Finally for all the $4$ clustering frameworks, the final values of hyper-parameters are: $k_{FA+SC} = 7$, $k_{PCA+SC} = 7$, $k_{FA+KMC} = 7$, and $k_{PCA+KMC} = 7$ respectively. Figure~\ref{fig:clusteringHP} shows the results of gap statistics for $2$ of these.

\section{Results \& Discussions}
\label{sec:results}
Different clustering frameworks produce different results and now the most crucial task is to compare these results and determine which framework generates best clusters. Section~\ref{sec:objectiveValidation} defines the proposed objective validation strategy which is cross-verified by subjectively validating the clustering results manually in Section~\ref{sec:subjectiveValidation}.

\subsection{Objective Validation}
\label{sec:objectiveValidation}
The proposed strategy for objective validation is based on the fact that in most tasks related to the clustering of residential electric demand profiles, pre-processing is done to define each household's daily consumption trend as a single vector of $24/r$ dimensions (or, in matrix representation, $\bm{M'_{n\times(24/r)}}$. In this case, $r=1$, $n=27$, and the pre-processing algorithm is the one described earlier in Algorithm~\ref{algo:preProcess}. If we randomly shuffle and split the raw data for each household (days$\times 24/r$) into $p$ partitions (\textit{w.r.t.} days) and implement the same pre-processing algorithm, we will obtain $p$ trends for each household instead of $1$. Now, assuming that:
\begin{itemize}
    \item $p \lll\textrm{data available for any one household}$, and
    \item pre-processing algorithm provides a reliable representation of the residential electric demand profiles,
\end{itemize}
we can argue that the resulting $p$ new trends of a particular will also fall into the same cluster where it was falling into originally.

Given the fact that the dimensionality reduction algorithms are generally deterministic (or non-random) in nature, we can easily perform the same steps on the $p$ trends to reduce their dimensions. Now, most clustering algorithms provide `cluster centers' and a list of `assigned labels' as output. Using this knowledge, we can easily determine which cluster center is closest to each of the dimensionally reduced $p$ trends of every household. If a partitioned trend of a particular household falls in the same cluster with the unpartitioned trend, we will call it a `match', otherwise it will be a `mis-match'. This further implies that more `matches' signify better clustering. Additionally, the whole process is repeated $100$ times to diminish any random error introduced during random shuffling. Complete pseudo-code for the objective validation strategy is provided in the Algorithm~\ref{algo:eval}.

\begin{algorithm}
%\DontPrintSemicolon % Some LaTeX compilers require you to use \dontprintsemicolon instead
\SetKwRepeat{Do}{do}{while}
$\vars{p} \gets \text{number of partitions for each household}$\\
$dist(\cdot, \cdot) \gets \text{function to calculate Euclidean distance}$\\
\Comment{Output from clustering algorithm}
$\textbf{\vars{labels}} \gets \text{labels assigned to each household}$\\
$\textbf{\vars{C}$_{(k \times d')}$} \gets \text{cluster centers of each cluster}$\\
\textbf{Initialize:}\linebreak
$\vars{match}, \vars{misMatch}, \vars{counter} = 0$\;
\Repeat{$\vars{counter}<100$}{
\ForEach{household}{
    $\vars{D } \gets \text{data for each household (days$\times$24/r)}$\;
    $\vars{D'} \gets \text{randomly shuffled data by rows}$\;
    Make \vars{p} equal partitions from rows of \vars{D'}\;
    \Comment{Perform Pre-Processing steps}
    $\bm{M}_{\vars{p}}^{(\vars{p} \times (24/r))} \gets \text{new medians of } \vars{p} \text{ partitions}$\;
    $\bm{M}_{\vars{p}}^{\prime} \gets \ell_2\text{-Normalization(}\bm{M}_{\vars{p}}\text{, row-wise)}$\;
    \Comment{Do Dimensionality Reduction}
    $\bm{N}_{\vars{p}}^{(\vars{p} \times d')} \gets \text{dimReduce(}\bm{M}_{\vars{p}}^{\prime}\text{, } d'\text{)}$\;
    \ForEach{$partition \in \{\vars{1}\cdots\vars{p}\}$ \textbf{as} $part$}{
        \Comment{Find Closest Cluster}
        $\vars{CC} \gets \argmin_{i}(dist(\bm{N}_{\vars{p}}[part],\textbf{\vars{C}\,[i,\,:]}))$\;
        \eIf{$\vars{CC} == \textbf{\vars{labels}}\,[household]$}{
            $\vars{match++}$\;
            }{
            $\vars{misMatch++}$\;
            }
        }
    }
    $\vars{counter++}$\;
}
$\vars{avgMatches}=\vars{match}/100$\;
$\vars{avgMisMatches}=\vars{misMatch}/100$\;
\KwResult{\vars{avgMatches} \& \vars{avgMisMatches}}
\caption{Objective Validation Strategy}
\label{algo:eval}
\end{algorithm}

It must be noted here that the complete clustering framework is being considered in the proposed objective validation strategy and not only the final clustering algorithm, leading to a robust comparison between the frameworks.

To comply with the condition on the values of $p$, the experiments were conducted only for $p=2$ and $p=3$ as the data was limited in the dataset used for this study. In Table~\ref{scoretable}, `\%Matches' are calculated by the following equation:
\begin{equation*}
    \text{\%Matches} = (\vars{avgMatches}\times100)/(n\times \vars{p})
\end{equation*}
where the values of $\vars{p}$ and $\vars{avgMatches}$ are the same as described in the Algorithm~\ref{algo:eval}. Similarly, `\%Mis-Matches' were calculated and reported alongside. Higher the value of `\%Matches' means better the performance of the framework. As shown in the Table~\ref{scoretable}, PCA$+$KMC frameworks performs the best among the $4$ considered clustering frameworks with `\%Matches'$=76.28$ for $p=2$, and `\%Matches'$=67.15$ for $p=3$.

\begin{table}[htb!]
\small
\centering
\caption{Results obtained by performing objective validation of the $4$ clustering frameworks.}
\begin{tabular}{ |lr||c|c| }
\hline
& Clustering Framework & \%Matches & \%Mis-Matches \\ 
\hline\hline
\parbox[t]{3mm}{\multirow{4}{*}{\rotatebox[origin=c]{90}{\small \textbf{p = 2}}}} 
& FA $+$ SC\ & 22.67 & 77.33\\ 
& FA $+$ KMC\ & 29.07 & 70.93 \\ 
& PCA $+$ SC\ & 18.78 & 81.22\\ 
& PCA $+$ KMC\ & \textbf{76.28} & 23.72\\
\hline\hline
\parbox[t]{3mm}{\multirow{4}{*}{\rotatebox[origin=c]{90}{\small \textbf{p = 3}}}} 
& FA $+$ SC\ & 21.34 & 78.66\\ 
& FA $+$ KMC\ & 24.98 & 75.02\\ 
& PCA $+$ SC\ & 17.60 & 82.40\\ 
& PCA $+$ KMC\ & \textbf{67.15} & 32.85\\
\hline
\end{tabular}

\vspace{-0.5cm}
\label{scoretable}
\end{table}

\subsection{Subjective Validation}\label{sec:subjectiveValidation}
\begin{figure*}[!htbp]
\begin{subfigure}{.49\textwidth}
  \centering
  % include first image
  \includegraphics[width=.75\linewidth]{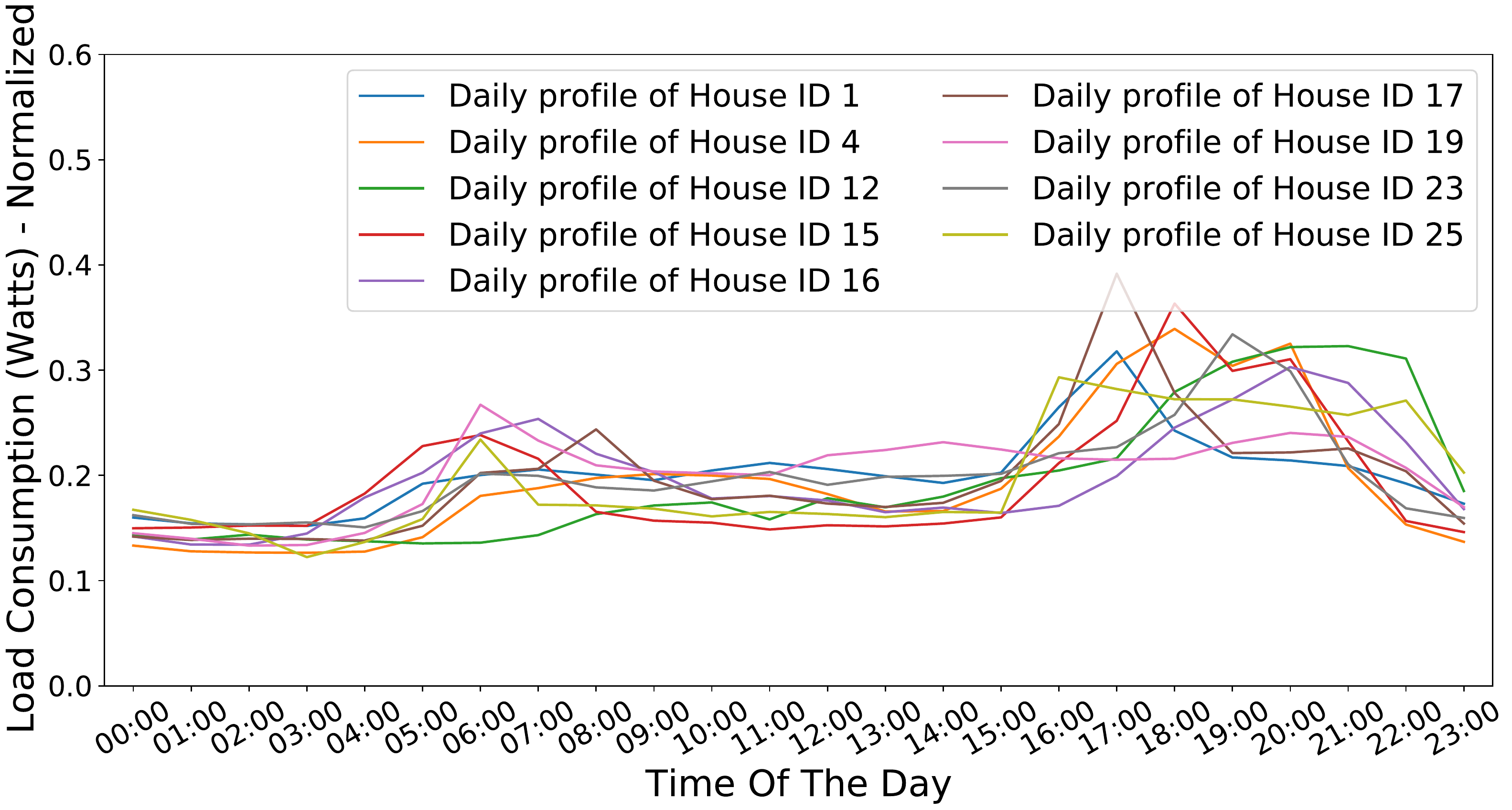}  
  \caption{Cluster $1$ as generated by the PCA$+$KMC framework.}
  \label{fig:cluster0}
\end{subfigure}\vspace{4pt}
\begin{subfigure}{.49\textwidth}
  \centering
  % include second image
  \includegraphics[width=.75\linewidth]{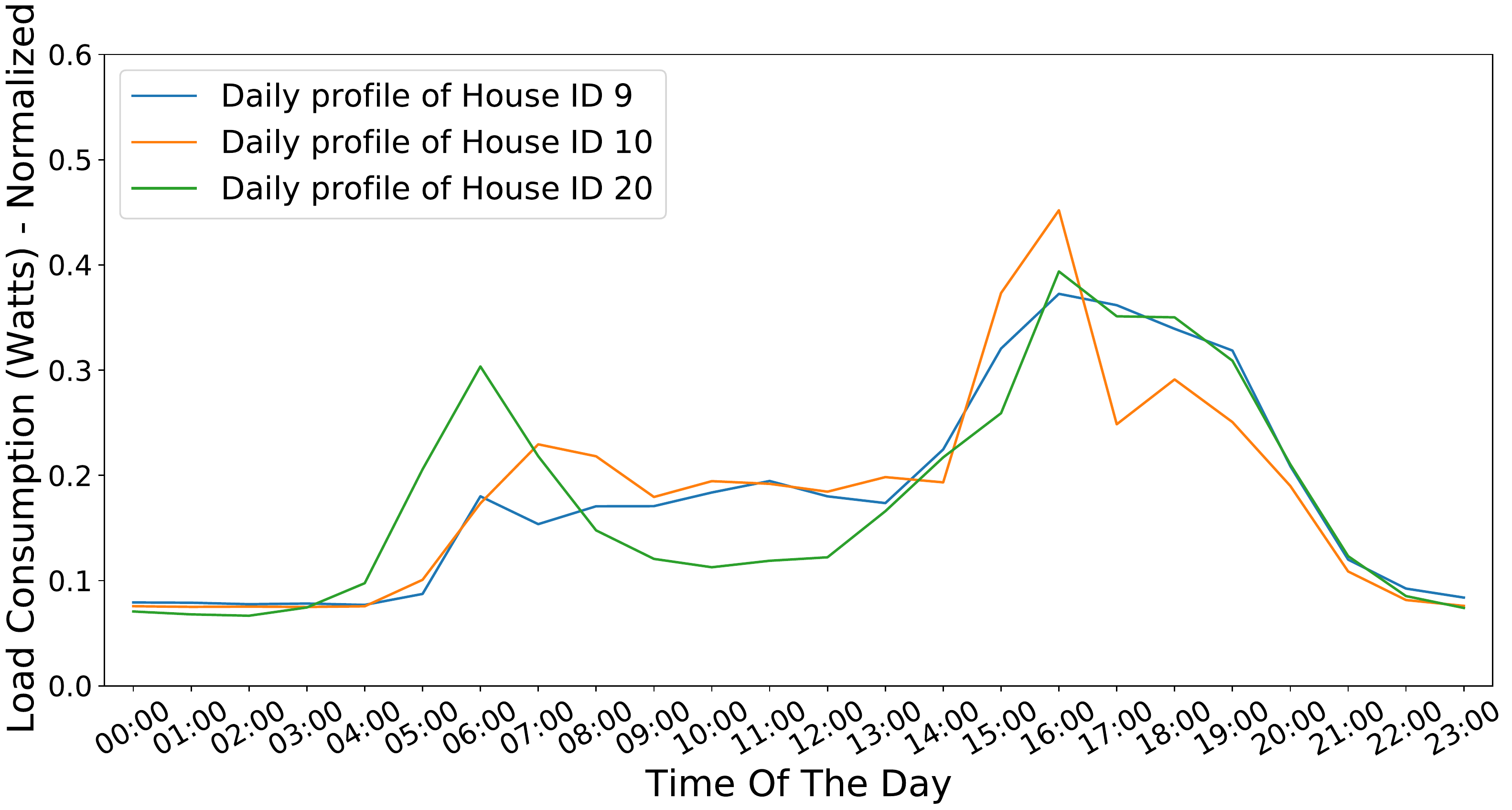}  
  \caption{Cluster $2$ as generated by the PCA$+$KMC framework.}
  \label{fig:cluster1}
\end{subfigure}
\newline
\begin{subfigure}{.49\textwidth}
  \centering
  % include third image
  \includegraphics[width=.75\linewidth]{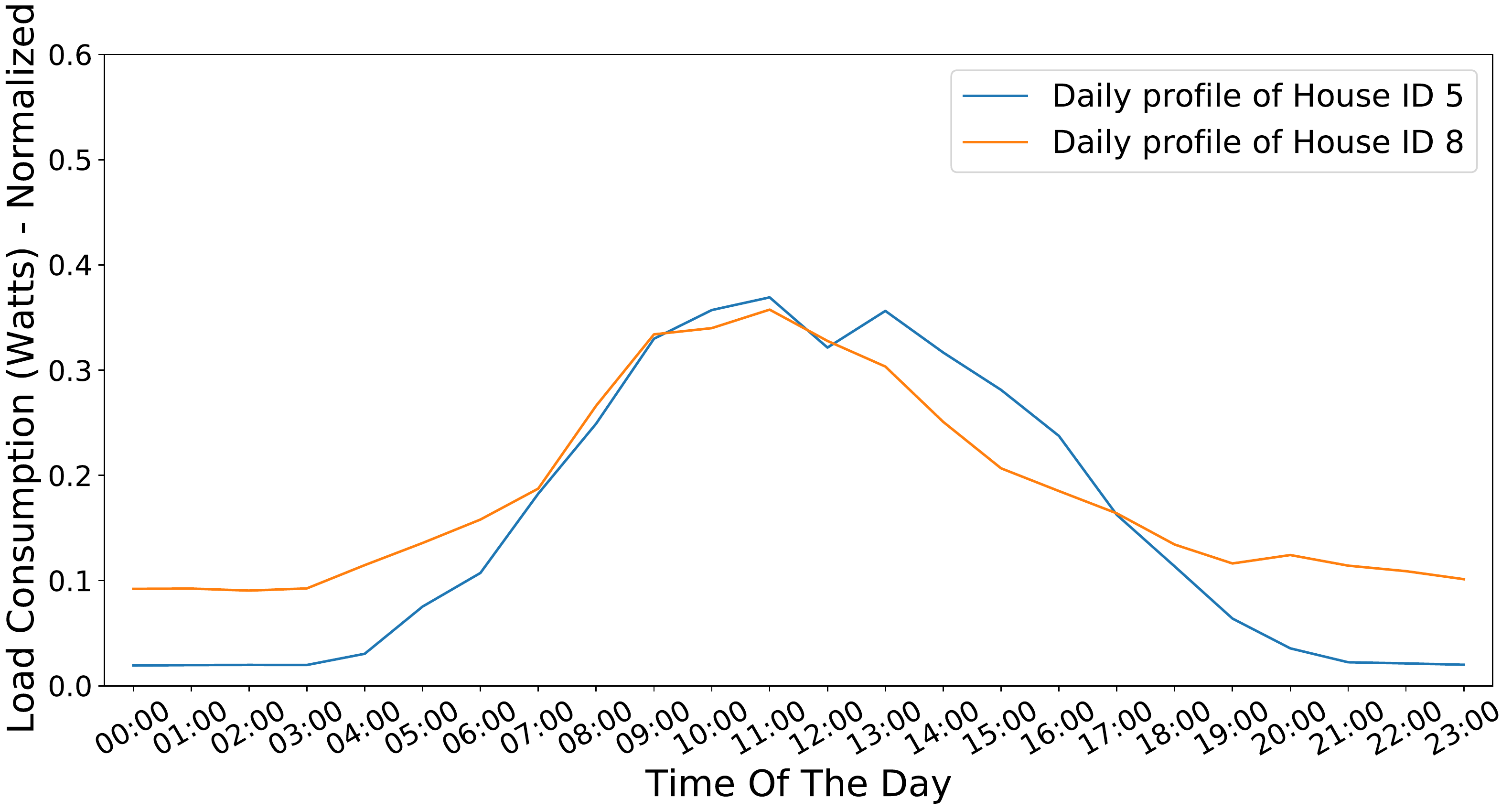}  
  \caption{Cluster $3$ as generated by the PCA$+$KMC framework.}
  \label{fig:cluster2}
\end{subfigure}\vspace{4pt}
\begin{subfigure}{.49\textwidth}
  \centering
  % include fourth image
  \includegraphics[width=.75\linewidth]{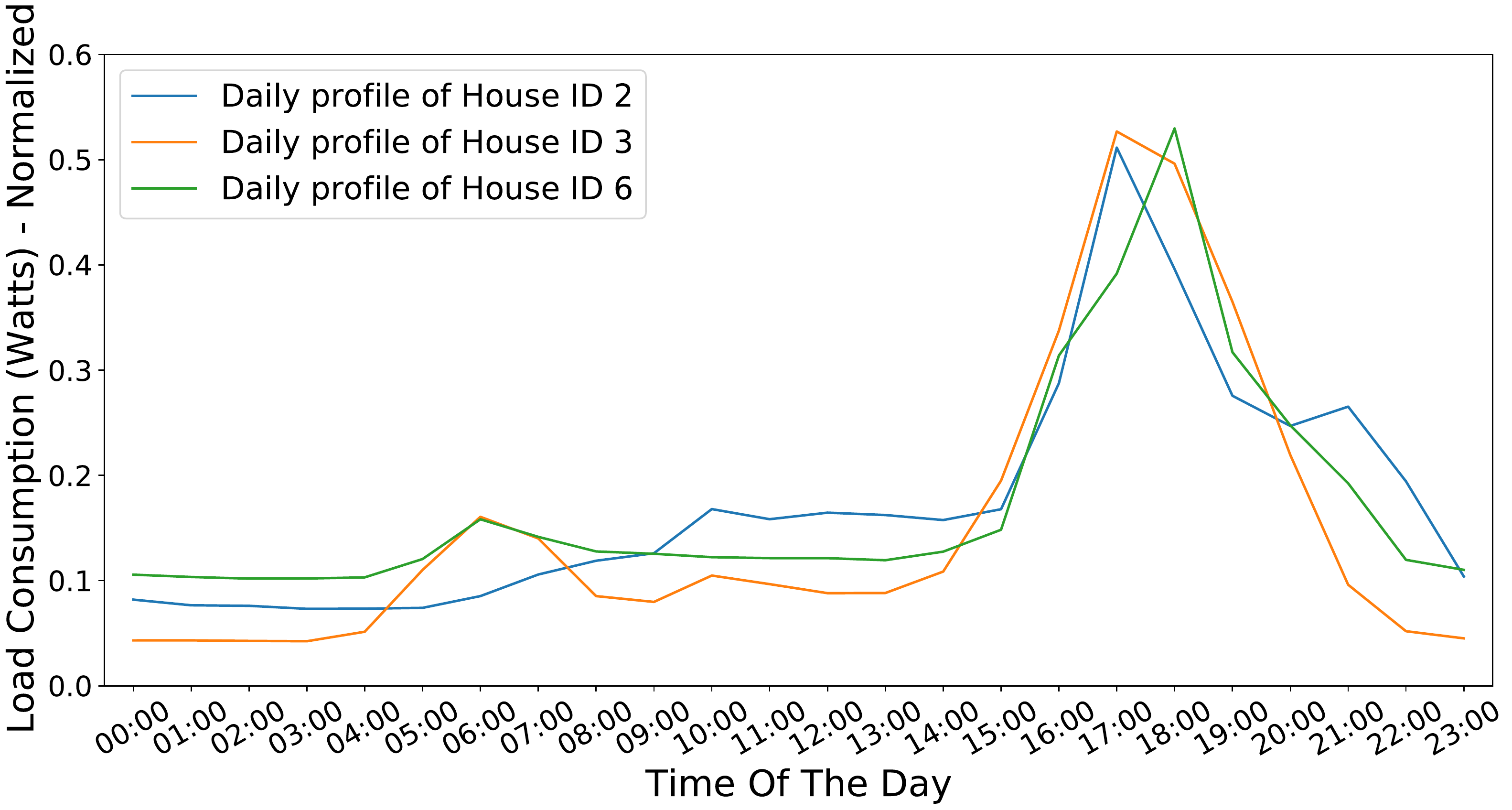}  
  \caption{Cluster $4$ as generated by the PCA$+$KMC framework.}
  \label{fig:cluster3}
\end{subfigure}
\newline
\begin{subfigure}{.49\textwidth}
  \centering
  % include fifth image
  \includegraphics[width=.75\linewidth]{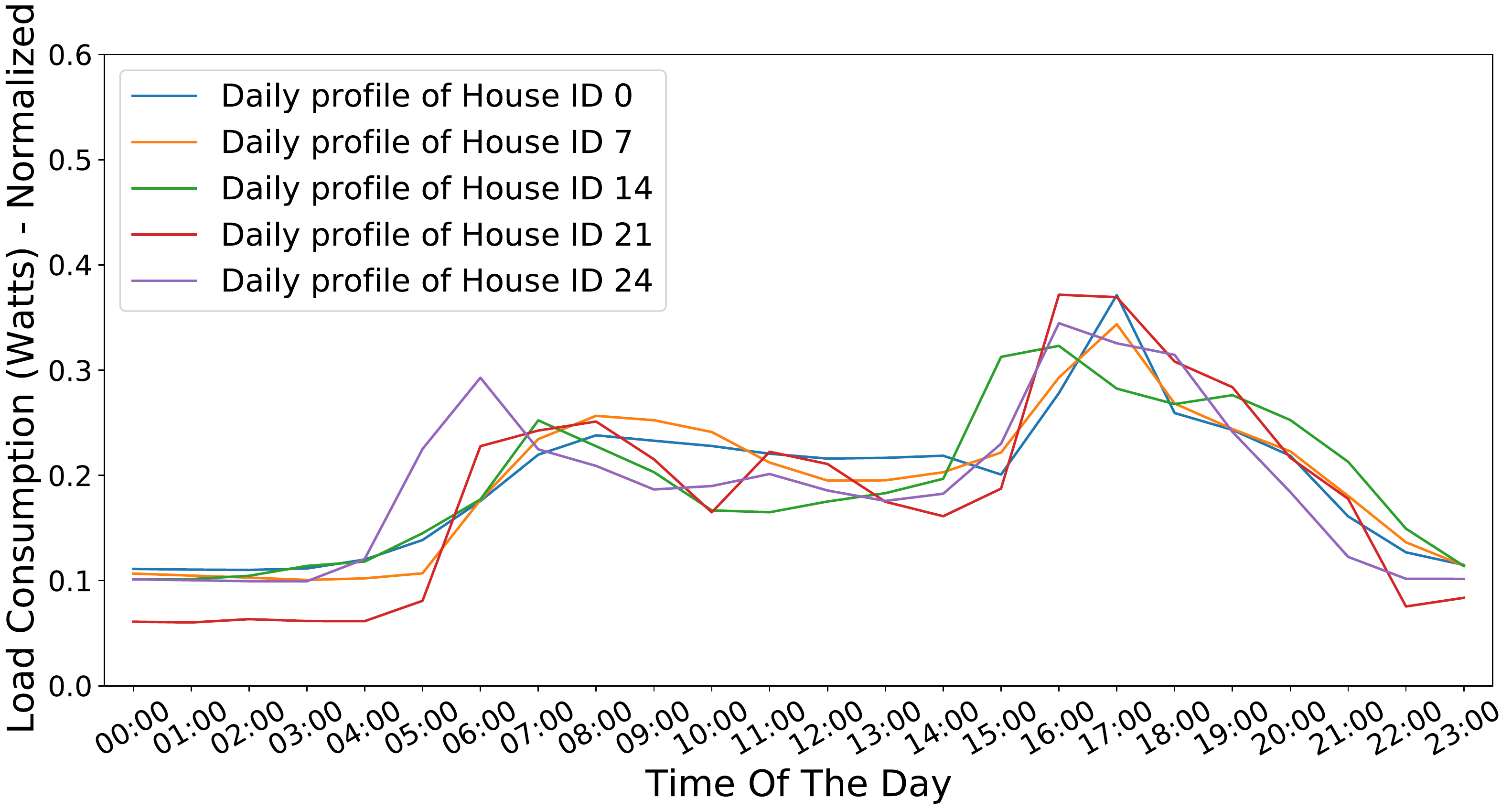}  
  \caption{Cluster $5$ as generated by the PCA$+$KMC framework.}
  \label{fig:cluster4}
\end{subfigure}\vspace{4pt}
\begin{subfigure}{.49\textwidth}
  \centering
  % include sixth image
  \includegraphics[width=.75\linewidth]{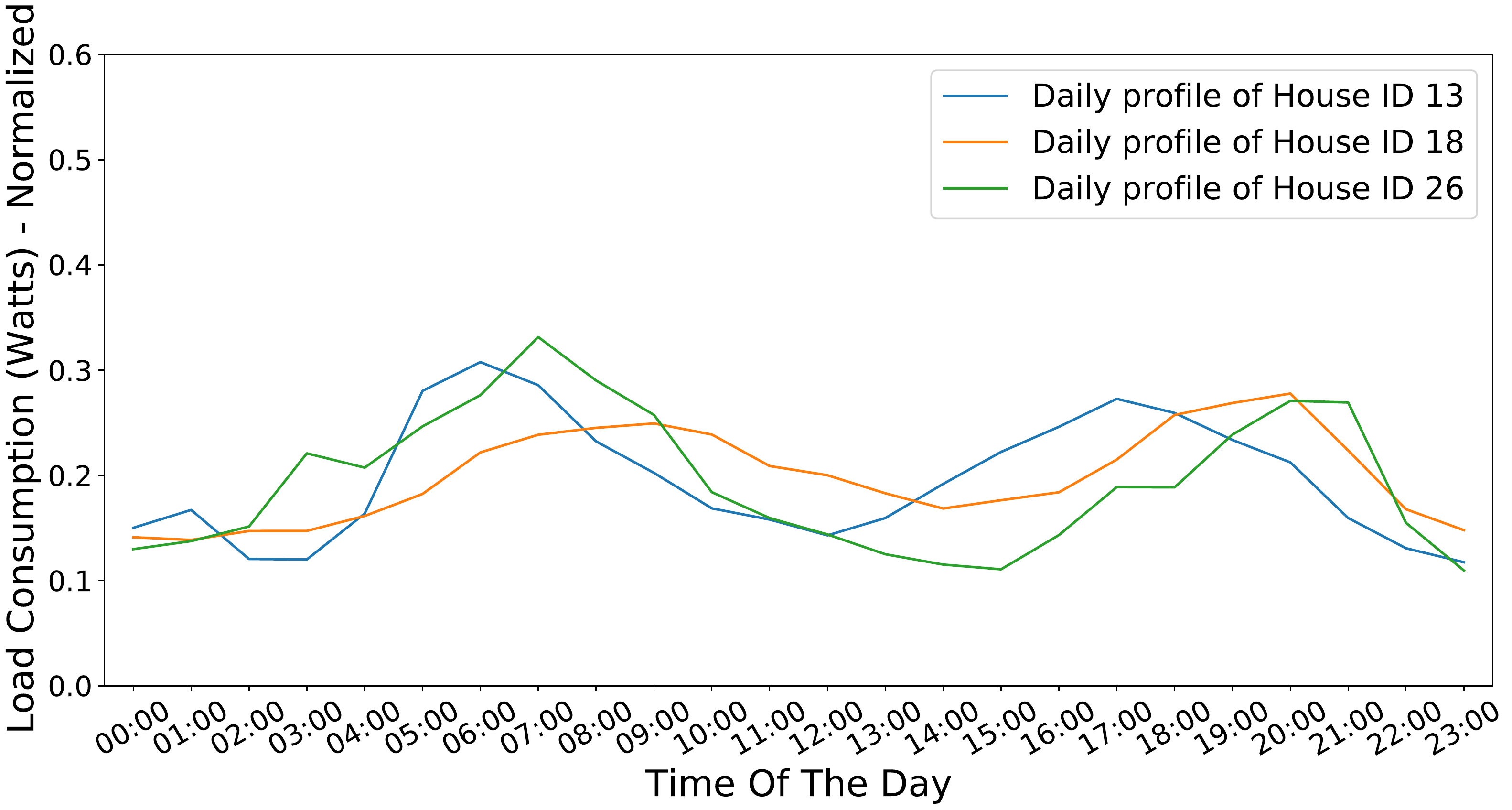}  
  \caption{Cluster $6$ as generated by the PCA$+$KMC framework.}
  \label{fig:cluster5}
\end{subfigure}
\newline
\begin{subfigure}{.49\textwidth}
  \centering
  % include seventh image
  \includegraphics[width=.75\linewidth]{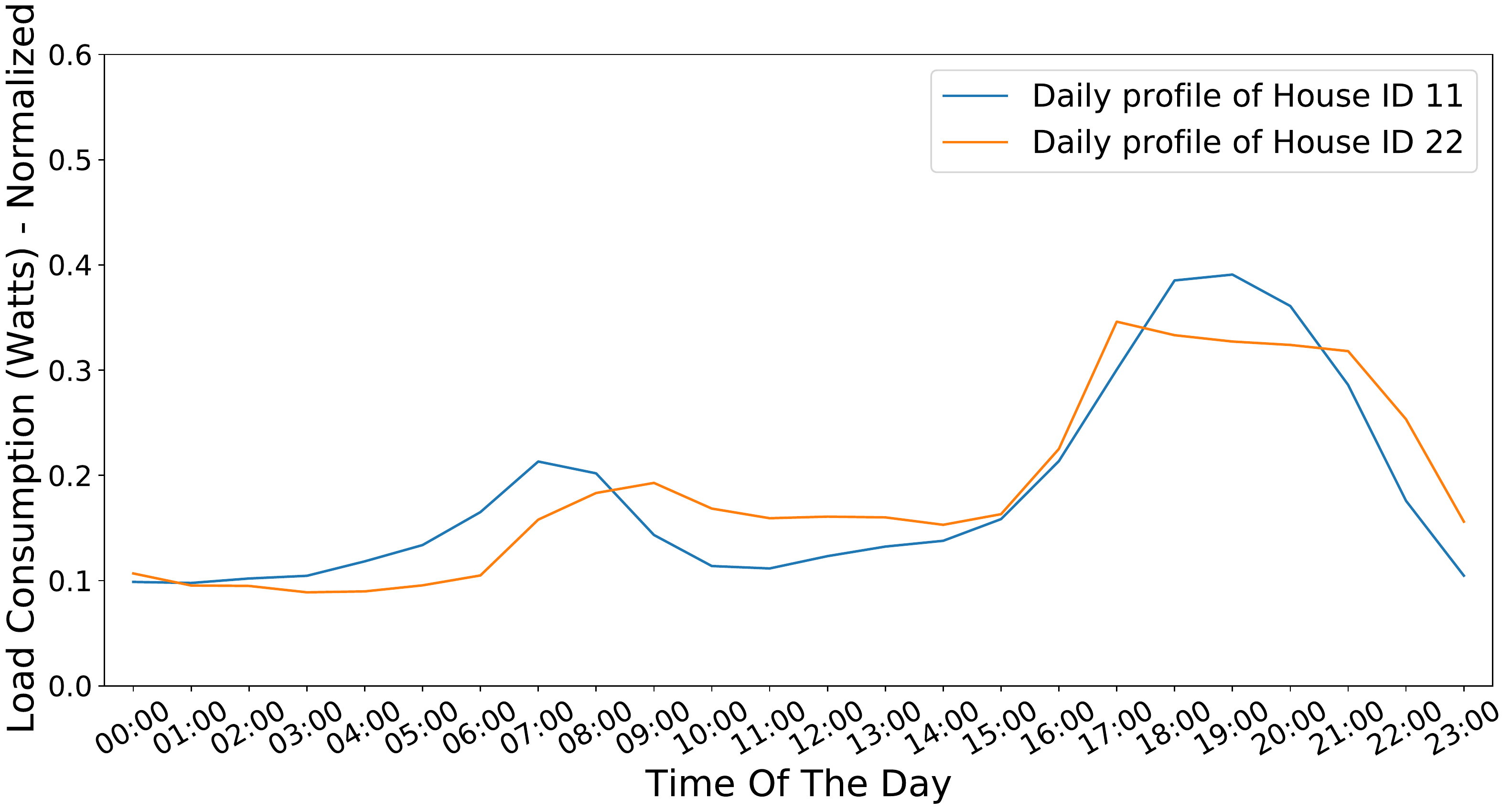}  
  \caption{Cluster $7$ as generated by the PCA$+$KMC framework.}
  \label{fig:cluster6}
\end{subfigure}\vspace{4pt}
\begin{subfigure}{.49\textwidth}
  \centering
  % include eighth image
  \includegraphics[width=.75\linewidth]{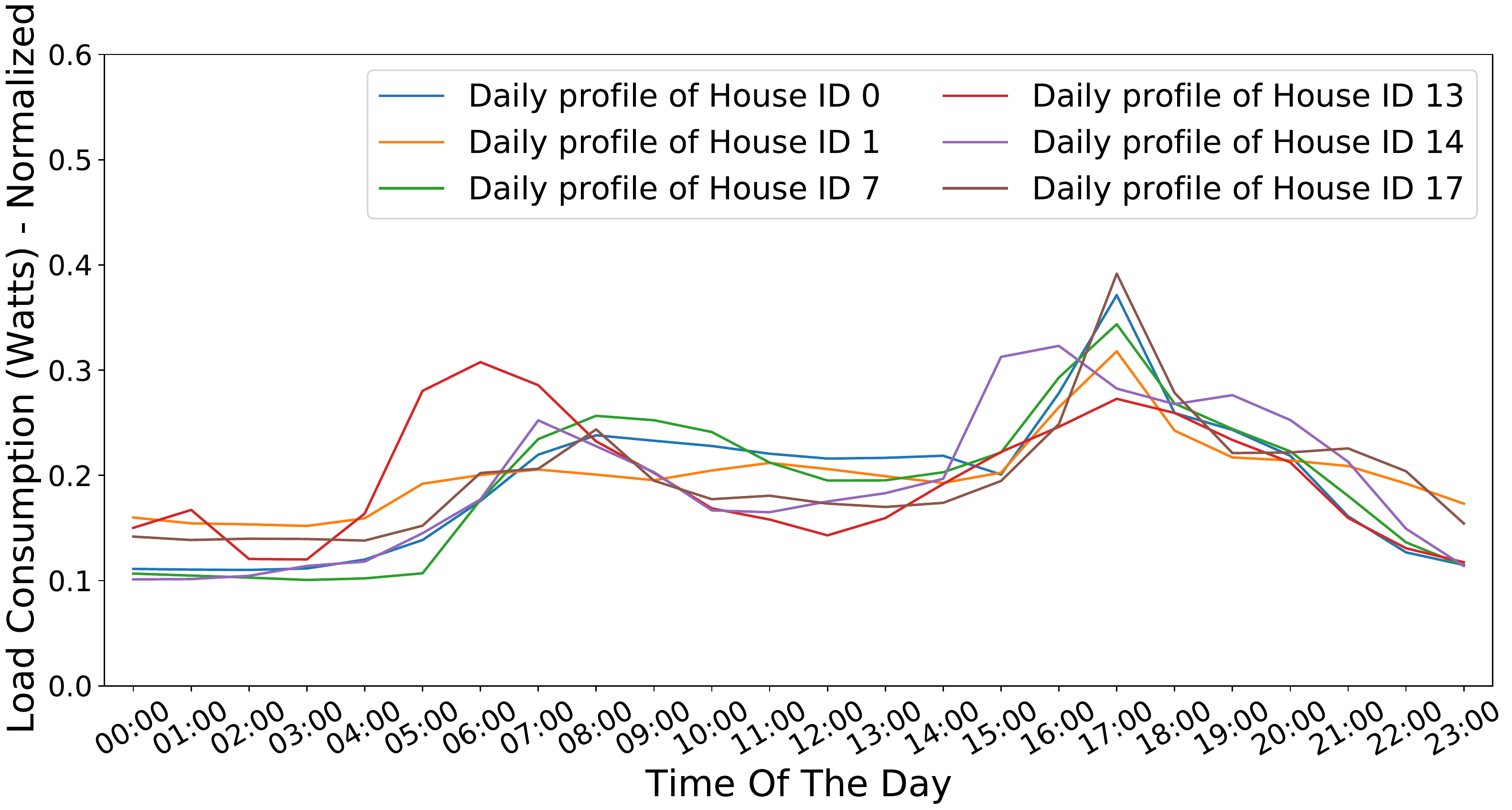}  
  \caption{Ill-defined cluster generated by the FA$+$KMC framework.}
  \label{fig:ILL-FAKM}
\end{subfigure}
\newline
\begin{subfigure}{.49\textwidth}
  \centering
  % include ninth image
  \includegraphics[width=.75\linewidth]{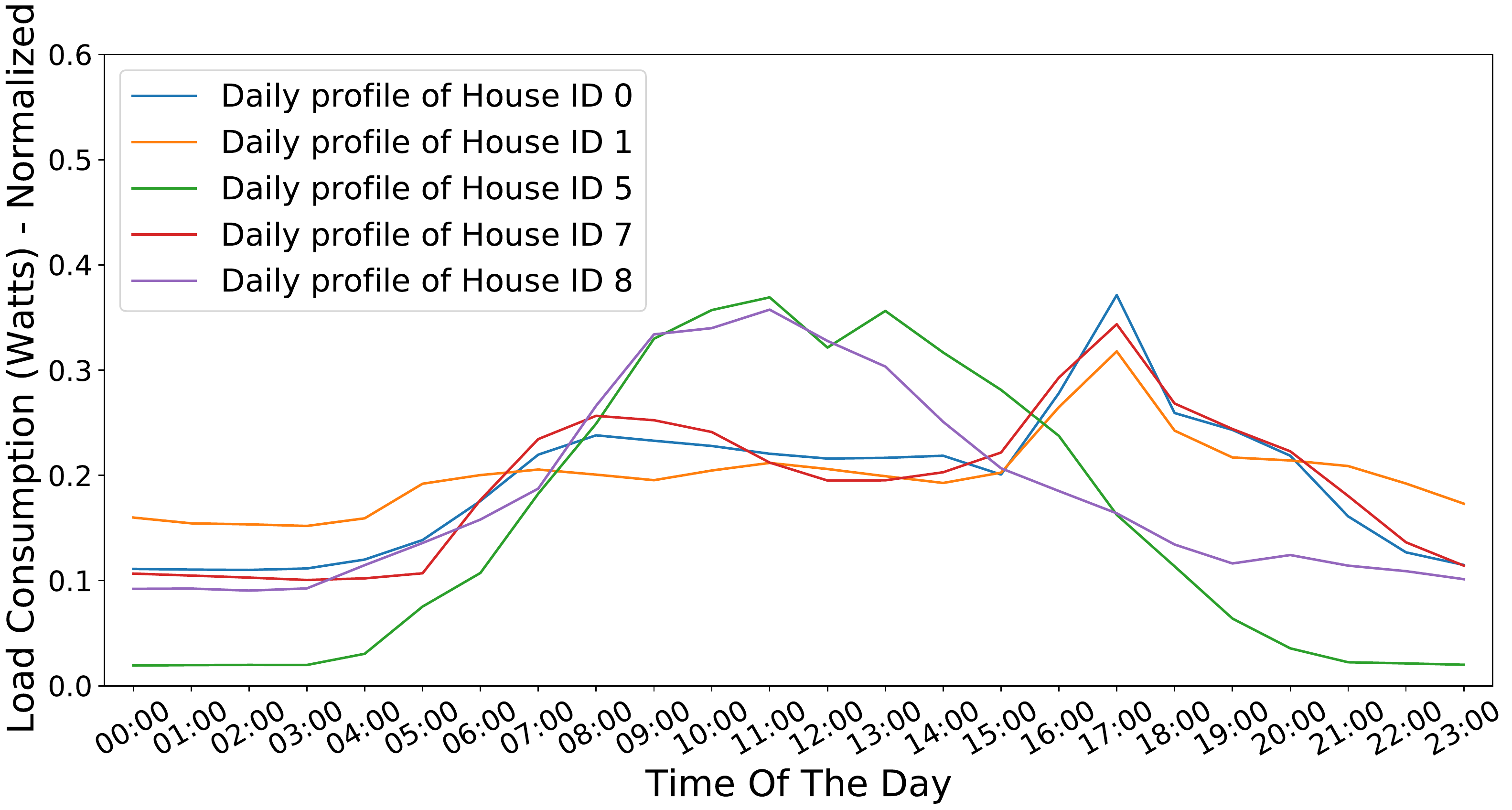}  
  \caption{Ill-defined cluster generated by the PCA$+$SC framework.}
  \label{fig:ILL-PCASC}
\end{subfigure}\vspace{4pt}
\begin{subfigure}{.49\textwidth}
  \centering
  % include tenth image
  \includegraphics[width=.75\linewidth]{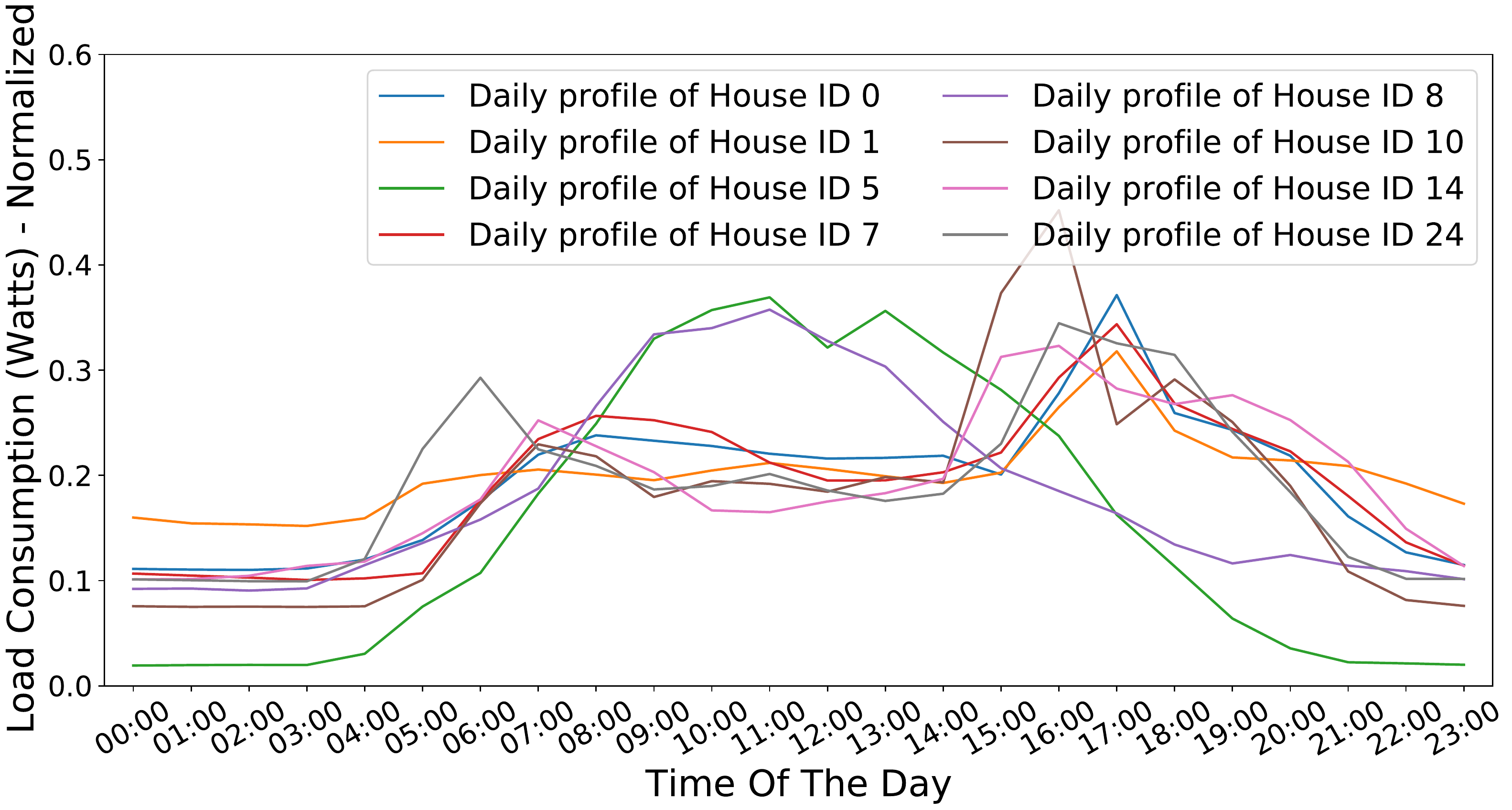}  
  \caption{Ill-defined cluster generated by the FA$+$SC framework.}
  \label{fig:ILL-FASC}
\end{subfigure}

\caption{Pre-processed daily electric demand profiles of households clustered by the clustering frameworks discussed in this work. Starting from the top-left and moving in line, the first $7$ sub-figures corresponds to the $7$ clusters generated by the `PCA$+$KMC' clustering framework. The last $3$ results are examples of defected clusters generated by the other $3$ frameworks, namely,  `FA$+$KMC' ,  `PCA$+$SC' , and  `FA$+$SC'.}
\label{fig:clusResults}
\end{figure*}

To cross-validate the results obtained from the objective validation strategy, manual inspection \cite{yildiz2018, Chicco2012} was carried out over the clustering results. Figure \ref{fig:clusResults} shows all the clusters generated by the recommended framework (i.e. PCA$+$KMC), alongside some examples of ill-defined clusters, one from each of the other frameworks. While all the households in each cluster plot for the proposed framework seem appropriately placed (see Fig.~\ref{fig:cluster0}--Fig.~\ref{fig:cluster6}), irregularities can easily be noticed for house ID $13$ in Fig.~\ref{fig:ILL-FAKM}, house ID $5$ and $8$ in Fig.~\ref{fig:ILL-PCASC}, and house ID $5$ and $8$ in Fig.~\ref{fig:ILL-FASC}. This further strengthens the claim that the recommendations made by the proposed objective validation strategy are reliable and robust not only to the clustering algorithm, but \textit{w.r.t} the complete framework.

\section{Conclusion and Future Work}
\label{sec:Conc}
In this paper, we have defined a clustering framework for identifying the consumption patterns of renewable energy in prosumers. Our case study is based on the data collected from $27$ households in Amsterdam, the Netherlands. Although the small data size leads to many difficulties in the clustering task, such instances of limited data availability are prominent in the real-world. The proposed objective validation strategy shows promise in reliably comparing the results of different clustering frameworks, even for such small datasets. In the future, we intend to analyze the impact of outliers on generated clusters, extend our study for a higher number of households, and with a larger statistical duration of data collection. This will assist us in relaxing the assumption that the behavior of each household is same throughout the year. We also plan to identify the impact of seasonal and demographic characteristics on the patterns of the electricity consumption, in addition to behavioral patterns. Lastly, we plan to include more dimensionality reduction and clustering algorithms, in future, which have been more recently reported to work effectively in electric load demand clustering tasks.	

\balance 


\end{document}